# Improving Candidate Generation
# for Low-resource Cross-lingual Entity Linking

**Shuyan Zhou, Shruti Rijhwani, John Wieting**
**Jaime Carbonell, Graham Neubig**

Language Technologies Institute
Carnegie Mellon University
{shuyanzh,srijhwan,jwieting,jgc,gneubig}@cs.cmu.edu

## Abstract

Cross-lingual entity linking (XEL) is the task of finding referents in a target-language knowledge base (KB) for mentions extracted from source-language texts. The first step of (X)EL is candidate generation, which retrieves a list of plausible candidate entities from the target-language KB for each mention. Approaches based on resources from Wikipedia have proven successful in the realm of relatively high-resource languages (HRL), but these do not extend well to low-resource languages (LRL) with few, if any, Wikipedia pages. Recently, transfer learning methods have been shown to reduce the demand for resources in the LRL by utilizing resources in closely-related languages, but the performance still lags far behind their high-resource counterparts. In this paper, we first assess the problems faced by current entity candidate generation methods for low-resource XEL, then propose three improvements that (1) reduce the disconnect between entity mentions and KB entries, and (2) improve the robustness of the model to low-resource scenarios. The methods are simple, but effective: we experiment with our approach on seven XEL datasets and find that they yield an average gain of 16.9% in TOP-30 gold candidate recall, compared to state-of-the-art baselines. Our improved model also yields an average gain of 7.9% in in-KB accuracy of end-to-end XEL.[1]

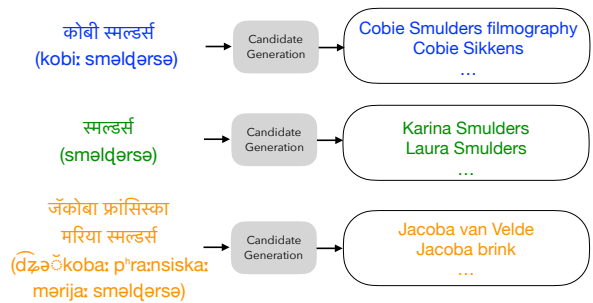

Figure 1: The candidate generation process for various mentions corresponding to the gold entity "Cobie Smulders". Strings on the left are mentions in the document, and the pronunciation in IPA of each string is written below it. The candidate entities in the English KB generated by the candidate generation model are shown on the right.

## 1 Introduction

Entity linking (EL; Bunescu and Paşca (2006); Cucerzan (2007); Dredze et al. (2010); Hoffart et al. (2011)) associates entity mentions in a document with their entries in a Knowledge Base (KB). In this work, we focus on cross-lingual entity linking (XEL; McNamee et al. (2011); Ji et al. (2015)) where the documents are in a source language that differs from the KB language (target). XEL is an important component task for information extraction in languages that do not have extensive KB resources, and can potentially benefit downstream applications such as cross-lingual building question answering systems (Veyseh, 2016), or supporting international humanitarian assistance efforts in areas that do not speak English (Strassel et al., 2017; Min et al., 2019).. Following Sil et al. (2018); Upadhyay et al. (2018a), we consider the target language KB to be English Wikipedia.

Given a document and named entity mentions identified by a Named Entity Recognition (NER) model, there are two primary steps in an XEL system: (1) *candidate generation*, in which a model retrieves a short list of plausible KB entities for each mention and (2) *disambiguation*, in which a model selects the most likely KB entity from the candidate list. The quality of candidate lists will influence the performance of the end-to-end XEL system, as correct entities not included in this list will not be recovered by the disambiguation model.

In monolingual EL, candidate generation has often been considered trivial (Shen et al., 2015). Simple approaches using string similarity or Wikipedia anchor-text links produce mention-entity lookup tables with high candidate recalls (e.g. in the 90% range), and thus most work focuses on methods for downstream entity disambiguation (Globerson et al., 2016; Yamada et al., 2017; Ganea and Hofmann, 2017; Sil et al., 2018; Radhakrishnan et al., 2018). String similarity (e.g. edit distance) cannot easily extend to XEL because surface forms of entities often differ significantly across the source and target language, particularly when the languages are in different scripts. Wikipedia link methods can be extended to XEL by using inter-language links between the two languages to redirect entities to the English KB (Spitkovsky and Chang, 2012; Sil and Florian, 2016; Sil et al., 2018; Upadhyay et al., 2018a). This method works to some extent, but often underperforms on low-resource languages due to the lack of source language Wikipedia resources.

---

[1]Code and data are avaliable at https://github.com/shuyanzhou/pbel_plus.

While scarce, there are some methods that propose to improve entity candidate generation by training translation models with LRL-English entity gazetteers (Pan et al., 2017), or learning neural string matching models based on an entity gazetteer in a related high-resource language (HRL) which is then applied to the LRL (Rijhwani et al., 2019) (more in §2). However, even with these relatively sophisticated methods, top-30 candidate still falls far behind their high-resource counterparts lagging by as much as 70% absolute candidate recall.

In this work, we perform a systematic study to understand and address the limitations of previous XEL candidate generation models. First, in §3 we examine the sources of error in the state-of-the-art candidate generation model of Rijhwani et al. (2019), and identify a number of potential reasons for failure. Specifically, we find that two common sources of error are (1) mismatch between the entity name in the KB and the entity mention in the text, and (2) failure of the string matching model itself. In Figure 1, we show an example of linking Marathi, a low-resource language spoken in Western India, to English, which we will use as a running example throughout the paper (although our method is broadly applicable, as noted in experiments). In this case, errors of the first type are due to the fact that the English entity *Cobie Smulders* is mentioned as स्मल्डर्स (green, Smulders) or जॅकोबा फ्रांसिस्का मरिया स्मल्डर्स (yellow, Jacoba Francisca Maria Smulders) in the text. Errors of the second type are simple recognition errors such as where the mention कोबी स्मल्डर्स (blue, Cobie Smulders) is recognized as English entity *Cobie Sikkens*. We proceed to propose methodological improvements that resolve these major issues.

The first set of improvements handles the mismatch between the unique entity name that appears in the English KB, and the many different realizations of it in the source text. First, we note that training data used in learning-based methods for XEL candidate generation (Pan et al., 2017; Rijhwani et al., 2019) is made of entity-entity pairs, which fail to capture this variation. We experiment with adding mention-entity pairs to the training data to provide explicit supervision, helping the model better capture the differences between mentions and entities (§4.1). Second, we note that many of the variations in the source language are actually similar to how the entity varies in English, and thus we can use English language resources to capture this variation. To this effect, we collect entity aliases from English Wikidata[2] and allow the model to also look up these aliases during the candidate generation process (§4.2).

The second contribution of this work is a better modeling strategy for strings that represent mentions and entities (§4.3). We posit that part of the reason why the LSTM-based model of Rijhwani et al. (2019) fails to properly model all words in a string is because it is not the ideal architecture to learn from limited training data, and as a result, it erroneously learns that some words in the mention can be ignored. To solve this problem, we replace the LSTM with a more direct model based on the sum of character $n$-gram embeddings (Wieting et al., 2016a), which we posit is more likely to generalize to this difficult learning setting.

We evaluate our proposed methods on four real world XEL datasets provided by DARPA LORELEI (Strassel and Tracey, 2016), as well as three other datasets we create with Wikipedia anchor-text and inter-language links (§5). While our methods are simple, they are highly effective – our proposed model leads to gains ranging from 7.4-33.3% in top-30 gold candidate recall compared to Rijhwani et al. (2019) in seven LRLs. Because our model provides downstream disambiguation models with a much larger head-room for improvement, we find that simply changing the candidate generation process yields an average gain of 7.9% in end-to-end XEL in-KB accuracy in four LRLs, pushing low-resource XEL a step towards high-resource XEL performance.

---

[2] https://www.wikidata.org/wiki/

## 2 Background

### 2.1 Problem Formulation

Given a set of mentions $\mathbf{M} = \{m_1, m_2, ..., m_N\}$ extracted from multiple documents in the source language, and an English KB $\mathcal{K}_{\text{EN}}$ that contains millions of entities with unique names, the goal of a candidate generation model is to retrieve a list of possible candidate entities $\mathbf{e}_i = \{e_{i,1}, e_{i,2}, ..., e_{i,n}\}$ from $\mathcal{K}_{\text{EN}}$ for each $m_i \in \mathbf{M}$. In consideration of the computational cost of the more complicated downstream disambiguation model, $n$ is often 30 or smaller (Sil et al., 2018; Upadhyay et al., 2018a). The performance of candidate generation is measured by the gold candidate recall, which is the proportion of retrieved candidate lists that contains the correct entity. It is critical that this number is high, as any time the correct entity is excluded, the disambiguation model will be unable to recover it. Formally, if we denote the correct entity of each mention $m$ as $\hat{e}$, the gold candidate recall $r$ is defined as:

$$r = \frac{\sum_{i=1}^{N} \delta(\hat{e}_i \in \mathbf{e}_i)}{N}$$

where $\delta(\cdot)$ is the indicator function which is 1 if true else 0, and $N$ is the total number of mentions among all documents. We follow (Yamada et al., 2017; Ganea and Hofmann, 2017) to ignore mentions whose linked entity does not exist in the KB in this work.[3]

We use "EN" to denote the target language English, "HRL" to denote any high-resource language and "LRL" to denote any low-resource language. For example, $\mathcal{K}_{\text{HRL}}$ is a KB in an HRL (e.g. Spanish Wikipedia), $e_{\text{HRL}}$ is an entity in $\mathcal{K}_{\text{HRL}}$. Since our focus is on low-resource XEL, the source language is always an LRL. We also refer to the HRL as the "pivoting" language below.

---

[3] The predictions of these mentions will always be wrong. This could be fixed by either designing mechanisms to predict "not linkable" or expanding the KB, which are beyond the scope of this work.

### 2.2 Baseline Candidate Generation Models

In this section, we introduce two existing categories of techniques for candidate generation.

**Direct Wikipedia-based Models** WIKIMENTION is a popular candidate generation model used by most state-of-the-art work in XEL (Sil and Florian, 2016; Sil et al., 2018; Upadhyay et al., 2018a). Specifically, this model first extracts a monolingual $m_{\text{LRL}}$-$e_{\text{LRL}}$ map from anchor-text links. For instance, if mention स्मल्डर्स (Smulders) is linked to entity कोबी स्मल्डर्स (Cobie Smulders) in some Marathi Wikipedia pages, कोबी स्मल्डर्स will be treated as a candidate entity of स्मल्डर्स. These Marathi entities are then redirected to their English counterpart by Wikipedia LRL-English inter-language links. For example, कोबी स्मल्डर्स (Cobie Smulders) will be redirected to *Cobie Smulders*. However, the reliance on the coverage of LRL Wikipedia strongly constrains this method in low-resource settings.

TRANSLATION is another Wikipedia based candidate generation model proposed by Pan et al. (2017). Instead of building a monolingual map that requires accessing anchor-text links in an LRL Wikipedia, this model translates any $m_{\text{LRL}}$ to $m_{\text{EN}}$ word-by-word and retrieves candidate entities from an existing $m_{\text{EN}} - e_{\text{EN}}$ map. The word-by-word translations are induced by LRL-English inter-language links. Even though TRANSLATION is less sensitive to the availability of resources (to some extent), its dependency on LRL-English inter-language links still limits its performance in low-resource settings.

**Pivoting-based Entity Linking** Instead of relying on LRL resources, pivoting-based entity linking (PBEL, Rijhwani et al. (2019)) learns to perform cross-lingual string matching based on an entity gazetteer between a related HRL and English. This model consists of two BI-LSTMs, namely, the HL-BI-LSTM and the EN-BI-LSTM. The training data is a collection of entity pairs ($e_{\text{HRL}} - e_{\text{EN}}$). Each of the BI-LSTMs reads in an entity name $e_{\text{HRL}}$ ($e_{\text{EN}}$) and encodes it to an embedding $\mathbf{v}_{\text{HRL}}$ ($\mathbf{v}_{\text{EN}}$). The learning objective is to maximize the similarity between the two entities of each pair. The trained model HRL is used as-is to encode the LRL mentions to $\mathbf{v}_{\text{LRL}}$, relying on the similarity between the languages to achieve a reasonably accurate encoding. A $\mathbf{v}_{\text{LRL}}$ is compared with every entity embedding in $\mathcal{K}_{\text{EN}}$, and entities with the

top-$n$ highest similarity scores are retrieved as the candidate entities. To compensate for the accuracy degradation due to transfer, this work also considers the similarity between $m_{\text{LRL}}$ and $e_{\text{HRL}}$, where $e_{\text{HRL}}$ is the counterpart of $e_{\text{EN}}$ in $\mathcal{K}_{\text{HRL}}$. Thus, the score between $m_{\text{LRL}}$ and entity $e_{\text{EN}}$ is defined as:

$$\text{score}(m_{\text{LRL}}, e_{\text{EN}}) = \max(\text{sim}(m_{\text{LRL}}, e_{\text{EN}}), \\ \text{sim}(m_{\text{LRL}}, e_{\text{HRL}})) \quad (1)$$

where $\text{sim}(x, y) = \text{cosine}(\mathbf{v}_x, \mathbf{v}_y)$. When $e_{\text{HRL}}$ does not exist, $\text{sim}(m_{\text{LRL}}, e_{\text{HRL}})$ is set to $-\infty$.

PBEL removes the reliance on LRL resources, and currently represents the state-of-the-art for candidate generation in low-resource XEL. However, as we analyze in detail in the following §3, it still faces a number of challenges.

## 3 Failures of Existing Models

In this section, we perform a systematic analysis of failure cases existing in PBEL (§3.1), and specifically focus on two error types: entity-mention mismatch (§3.2) and string matching failures (§3.3).

### 3.1 Mention Types and Analysis

We apply a PBEL model trained with $e_{\text{HRL}} - e_{\text{EN}}$ pairs to generate candidate entities for mentions extracted from LRL documents. For LRLs we use Tigrinya, Oromo, Marathi and Lao, and for HRLs we use Amharic, Hindi, Hindi, and Thai respectively. The details of the datasets are in §5. We randomly sample 100 system outputs from each LRL and manually annotate their mention type according to an typology created simultaneously while performing analysis. The mention type is as follows, where the comparison is between the mention in a LRL and the entity string in English:

**DIRECT:** The mention is a direct transliteration of the entity. For example, one a mention of *Cobie Smulders* is कोबी स्मल्डर्स (Cobie Smulders)

**ALIAS:** The mention is another full *proper* name that is different from the entity name in English KB. For instance, a mention of *Cobie Smulders* as जॅकोबा फ्रांसिस्का मरिया स्मल्डर्स (Jacoba Francisca Maria Smulders).

**TRANS:** The mention and the entity have word-by-word alignment, however, the mention contains regular words (e.g. university, union) that cannot be transliterated directly.

**EXTRA_SRC:** There is at least one extra word in the mention that is *not* a proper noun (e.g. श्री (Sir)); or there is at least one extra syllable in the mention, which is often due to the morphology of the source language.

**EXTRA_ENG:** There is at least one extra word in the English entity that is *not* a proper noun.

**BAD_SPAN** : The mention span is not an entity due to mis-annotation, or non-standard anchor text in Wikipedia; the annotated linked entity is wrong; the mention is in another language other than our testing language.

We consider three situations for each sample: (1) in top-1: the model ranks the correct entity the highest, the ideal case; (2) in top-2 to 30: the model ranks the correct entity in the top-2 to top-30, which is less ideal, but will still potentially allow a downstream disambiguation model to predict the correct entity and (3) not in top-30: the model does not rank the entity to top-30, which will certainly lead to an error.

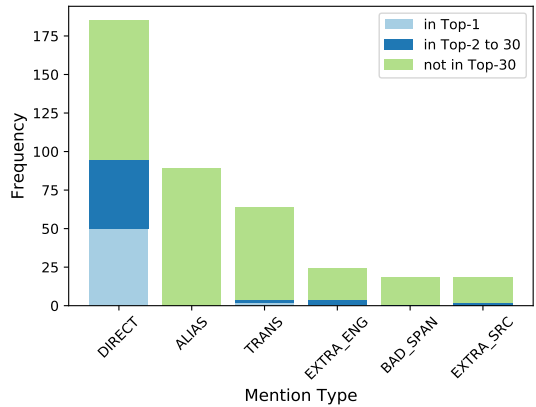

Figure 2: The distribution of mention types in 400 samples and the baseline model's performance with respect to each of the mention types.

Figure 2 shows the mention types of the 400 samples and PBEL performance within each of the mention types. In the following sections, we examine, in depth, two major causes of error: mention-entity mismatch (largely affecting errors in ALIAS, EXTRA_SRC, and EXTRA_ENG categories), and model failure (largely affecting errors in DIRECT).

### 3.2 Failures due to Mention-Entity Mismatch

As demonstrated in Figure 1, a single English entity can have different realizations in the source language document. As a result, many of these realizations will not match lexically or phonetically with the entity in the KB. This poses a serious problem for matching methods that rely on graphemic or phonemic similarity such as PBEL.

| Lang | am | so | hi | th |
|---|---|---|---|---|
| $|e_{\text{HRL}}|=|e_{\text{EN}}|$ | 82.9 | 80.7 | 83.4 | 56.8 |
| $|m_{\text{HRL}}|=|e_{\text{EN}}|$ | 71.1 | 58.0 | 56.8 | 55.8 |

Table 1: Proportion of entries where HRL strings have the same number of words as their English counterparts.

One typical pattern in mention-entity variation is additional words, as noted in the EXTRA_SRC and EXTRA_ENG classes. We examine more systematically across the whole corpus by comparing the number of words on each side, which is a rough lower bound on the amount of this mismatch. The first row in Table 1 is the comparison between $e_{\text{HRL}}$ and $e_{\text{EN}}$, which presumably have better word-by-word alignment (and were used in training of previous XEL methods). The second row displays the comparison between $m_{\text{HRL}}$ and $e_{\text{EN}}$. It is obvious that entity-entity pairs have more consistent length in words, while this consistency is not preserved in mention-entity pair data. Thus, even if the previous PBEL model could easily learn exact string matches from the entity-entity training data, to successfully associate mention-entity pairs, the model would need to capture more complex patterns (e.g., ignoring some words).[4]

The diverse realizations of a single entity bring another, more serious, challenge to models that mainly learn string matches: in reality, a realization does not necessarily have significant overlap with the entity name in Wikipedia. Sometimes, the mention does not have any overlap with the entity name at all, as noted in the ALIAS class. This common pattern reflects the limitation of using $e_{\text{EN}}$ as the unique representation on the English side.

### 3.3 Failures in Direct Transliteration

Even in seemingly easy cases where the entity is a perfectly transliteration of the mention (DIRECT), we found the LSTM to fail frequently in our low-data scenario. Among all DIRECT errors, we found an interesting observation that the BiLSTM often only properly captures the first word (or the first a few characters) and ignores the existence of the second and further-on words. For example, the model ranks *Cobie Sikken* higher than *Cobie Smulders* for कोबी स्मल्डर्स (Cobie Smulders).

---
[4]Low numbers for *th* are due to lack of explicit word boundaries marked by spaces.

To better understand this behavior, we manually annotated 100 training pairs in Hindi and measured how often the second or later words in $e_{\text{HRL}}$ do not match their counterpart in $e_{\text{EN}}$ *phonologically*.[5]

We find that while 93 examples share a phonologically similar first word, about 40 of them have second and further-on words that are not phonological matches: while most pairs have word-by-word mappings, their second or later words often match with each other only *semantically* – i.e. there are regular words (e.g. district, university) that have very different pronunciations across the HRL and English, and are therefore difficult to predict unless they are explicitly seen in the training data. The BiLSTM, which is a flexible model, seems to overfit and erroneously learn that latter words in the sentence do not need to be mapped directly with little inductive bias. This is a straightforward explanation for why the model learns to ignore the second and further-on words.

To sum up, the failures of the PBEL model can be mainly attributed to 1) lack of explicit supervision; 2) lack of external resources to assist cases where the mention and entity name diverge significantly and 3) the BiLSTM's inability to properly match the whole string.

## 4 Improved Candidate Generation

Based on the results of this empirical study, we propose three methods to resolve the main problems inherent in the baseline PBEL model.

### 4.1 Eliminating Train-Test Discrepancy

The mention-entity discrepancy naturally leads to our first simple but effective improvement to the baseline model: we extend the original $e_{\text{HRL}} - e_{\text{EN}}$ pairs with $m_{\text{HRL}} - e_{\text{EN}}$ pairs. We first collect $m_{\text{HRL}} - e_{\text{HRL}}$ pairs from anchor-text links in an HRL Wikipedia and then redirect these entities to their parallel in English Wikipedia. As a result, we get the desired $m_{\text{HRL}} - e_{\text{EN}}$ pairs. For instance, if स्मल्डर्स (Smulders) is linked to कोबी स्मल्डर्स (Cobie Smulders) in some Marathi Wikipedia pages, which could be redirected to Cobie Smulders in English, स्मल्डर्स and Cobie Smulders form one mention-entity pair. While this is perhaps obvious in hindsight, to our knowledge, all previous works that explicitly train XEL candidate retrieval models do so on $e_{\text{HRL}} - e_{\text{EN}}$ pairs (Pan et al., 2017; Rijhwani et al., 2019), which are mostly word-by-word mappings.

---
[5]The phonological similarity of names across languages is vital to the success of cross-lingual mention-entity matching.

### 4.2 Utilizing English Entity Aliases

The training method introduced in the previous section will render the model more capable of dealing with minor differences between mentions and entities. However, it still would struggle to match strings with significant differences, such as the examples of "Cobie Smulders" and "Pope Paul V" shown in Section 3.2. To mitigate this, we propose using Wikidata, a crowd-edited knowledge base similar to Wikipedia, which provides an "also known as" section that lists common aliases of each entity.[6] Our second method is based on the observation that Wikidata resources can serve as an off-the-shelf alias lookup table with better coverage than simply using the entity's canonical Wikipedia title. An example of how this lookup table can increase coverage is indicated in Figure 2. In our analysis, we found that more than 50% of the ALIAS mentions could be covered by this table. There is a map between Wikipedia entities and Wikidata entities, so we can direct Wikipedia to the Wikidata to retrieve these aliases.[7]

At test time, we treat the alias of an entity equally as its main Wikipedia entity name, allowing the model to match the target mention to this alias as well. As a result, $\text{sim}(m_{\text{LRL}}, e_{\text{EN}})$ in Equation (1) is modified as:

$$\text{sim}(m_{\text{LRL}}, e_{\text{EN}}) = \max_{a_i \in \mathbf{A}} \left( \text{sim}(m_{\text{LRL}}, a_i) \right)$$

where $\mathbf{A}$ is a combination of entity Wikipedia title and entity aliases.[8] Note that while one may consider using aliases in languages other than English, we found that they are very scarce, so we did not attempt to expand entity names on the HRL side.

### 4.3 More Explicit String Encoding

As mentioned previously, while BI-LSTMs have proven powerful in modeling sequential data in the literature, we argue that they are not an ideal string encoder for this setting. This is because our training data contains a nontrival number of pairs that contains less predictable word mappings (e.g. translations). With such large freedom in the face of insufficient and noisy training data, this encoder seemingly overfits, resulting in poor generalization. Previous work (Dai and Le, 2015; Wieting et al., 2016b) have noticed similar problems when using LSTMs for representation learning.

---
[6]e.g. https://www.wikidata.org/wiki/Q200566

[7]Other resources such as bold terms, link anchors, disambiguation pages and surnames of mentions could potentially increase the coverage of Wikidata.

[8]Note that incorporating aliases results in a small amount of extra computation by multiplying the effective size of the KB by $a$, the average number of aliases per mention. However, in Wikidata, $a = 1.2$, so we believe this is a reasonable cost-benefit trade-off, given the gains afforded by incorporating these aliases for many languages.

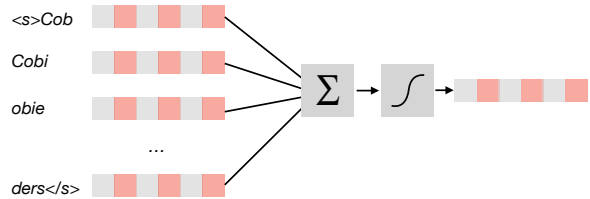

Figure 3: The architecture of CHARAGRAM.

As an alternative, we propose the use of the CHARAGRAM model (Wieting et al., 2016a) as the string encoder. This model scans the string with various window sizes and produces a bag of character $n$-grams. It then maps these $n$-grams to their corresponding embeddings through a lookup table. The final embedding of the string is the sum of all the $n$-gram embeddings followed by a nonlinear activation function. Figure 3 shows an illustration of the model.

Formally, we denote a string as a sequence of characters $\mathbf{x} = [x_1, x_2, ..., x_m]$ which includes space characters as well as special start and end symbols. We use $x_i^j$ to denote a sub-sequence from position $i$ to position $j$ inclusive. For example, $x_i^j = [x_i, x_{i+1}, ..., x_j]$. The embedding $\mathbf{v}$ of a string $\mathbf{x}$ is:

$$\mathbf{v} = \tanh \left( \mathbf{b} + \sum_{i=1}^{m} \sum_{n \in \mathbf{N}} \mathbb{1}(x_{i+1-n}^i \in V) W_{x_i^j} \right)$$

where $\mathbf{N}$ is a set of predefined window sizes. $\mathbf{b} \in \mathbb{R}^d$, $V$ is all $n$-grams seen in the training data, $W \in \mathbb{R}^{|V| \times d}$ is the embedding lookup table and $W_{x_i^j} \in \mathbb{R}^d$ is the embedding of $x_i^j$. Note that $\mathbb{1}(x)$ is the indicator function, if a $n$-gram is not in $V$, we simply discard it.

Compared to the BI-LSTM, the advantages of CHARAGRAM are four-fold. First, the complexity of memorizing short character strings in the model is reduced. CHARAGRAM learns multi-character subsequences by simply adding them to an embedding table, whereas the LSTM learns them in a multi-step recurrent process. Second, due to their relatively higher expressiveness, LSTMs overfit to the noisy and relatively small training data provided by Wikipedia bilingual entity maps, the likely reason for LSTMs only considering the start word in errors from the DIRECT category. In contrast, CHARAGRAM does not consider order information, giving it an explicit inductive bias that forces it to rely on character $n$-gram matching for all $n$-grams in the sequence. Third, CHARAGRAM's simple architecture eases the learning process. For instance, the LSTMs needs $O(m)$ steps to propagate gradients from start to finish (Vaswani et al., 2017), while the CHARAGRAM requires only $O(1)$ step to do so. Finally, while not a performance-based advantage, the CHARAGRAM model is more interpretable, which make our further analysis easier to perform (see Section 5).

We follow Wieting et al. (2016a); Rijhwani et al. (2019) and use negative sampling with a max-margin loss to train the model:

$$L = \sum_{i=1}^{B} \max(0, 1 - \text{sim}(m, e_{\text{EN}^+}) + \text{sim}(m, e_{\text{EN}^-}^i))$$

where $e_{\text{EN}^+}$ is the linked entity of $m$ and $e_{\text{EN}^-}$ is a randomly sampled English entity. $B$ is the number of negative samples for each positive pair.

## 5 Experiments

### 5.1 Datasets

We evaluate our model on the following datasets, spanning seven low-resource languages.

**DARPA-LRL:** The data for the first four languages are news articles, blogs, and social media annotated with entity spans and links by LDC as part of the DARPA LORELEI[3] program. The documents are in four low-resource languages: Tigrinya (`ti`; a Semitic language spoken in Eritrea and Ethiopia, written in Ethiopian script), Oromo (`om`; an Afroasiatic langage spoken in the Horn of Africa, written in Roman script), Kinyarwanda (`rw`; a language of the Niger-Congo family spoken in Rwanda, written in Roman script) and Sinhala (`si`, and Indo-Aryan language spoken in Sri Lanka, written in its own script). These are naturally-occurring real-world data annotated and linked to a KB, containing information about disasters and humanitarian crises. We use these as the "gold standard" datasets for our evaluation.

**WIKI:** One disadvantage of the DARPA-LRL dataset, however, is that it is not publicly distributed at the time of this writing. In order to allow for direct comparison with our method by researchers without access to the DARPA-LRL data, we additionally create three datasets from Wikipedia, as described in §4.1. Specifically, these include Marathi (`mr`, an Indo-Aryan language spoken in Western India, written in Devanagari script), Lao (`lo`, a Kra-Dai language written in Lao script) and Telugu (`te`, a Dravidian language spoken in southeastern India written in Telugu script). As Wikipedia is created through crowd-sourcing, the anchor-text links are similar to those appearing in realistic XEL datasets. It is notable that entity mentions in WIKI often closely match the Wikipedia entity titles, and thus this dataset is nominally easier than the DARPA-LRL dataset.

### 5.2 Training Details

In the CHARAGRAM model, we use character $n$-grams with $n \in \{2, 3, 4, 5\}$, and embedding size of 300. We train the model with stochastic gradient descent (SGD) with batch size 64, and a learning rate of 0.1. For the BI-LSTM model, we follow Rijhwani et al. (2019) for hyperparameter selection.

We also compare our model with a character-based CNN with sum-pooling (CHARCNN; (Zhang et al., 2015; Wieting et al., 2016a)), where parameters are set to be roughly comparable in size to our CHARAGRAM model. The embedding size of each character is set to 1024; the kernel size is set to 2, 3, 4, 5 each with 4800 feature maps. The output of sum-pooling layer with a dimension of 19200 (4800×4) is fed a fully connected layer and results in a vector of size 300. The dropout is set to 0.5.[9]

For each training language, we set aside a small subset of training data ($m_{\text{HRL}} - e_{\text{EN}}$) as our development set. For all models, we stop training if top-30 gold candidate recall on the development set does not increase for 50 epochs, and the maximum number of training epochs is set to 200.

We select the HRL that has the highest character $n$-gram overlap with the source LRL, a decision we discuss more in §5.4. Rijhwani et al. (2019)

---

[9] We also try smaller architectures with embedding size set to 64 and number of feature maps set to 300. This configuration yields worse performance than the larger model.

| LRL | HRL | Representation |
|-----|-----|----------------|
| `ti` | Amharic (`am`) | Phoneme |
| `om` | Indonesian (`id`) | Grapheme |
| `rw` | Tagalog (`tl`) | Phoneme |
| `si` | Hindi (`hi`) | Phoneme |
| `mr` | Hindi (`hi`) | Grapheme |
| `lo` | Thai (`th`) | Phoneme |
| `te` | Hindi (`hi`) | Phoneme |

Table 2: The HRL for each LRL. For phoneme representations, all input strings in LRL, HRL, and English are convert to IPA. For grapheme representations, strings preserve their original representation.

used phoneme-based representations to help deal with the fact that different languages use different scripts, and we do so as well using Epitran (Mortensen et al., 2018) to convert strings to international phonetic alphabet (IPA) symbols. The selection of the HRL and the representation of each LRL is shown in Table 2. Epitran has relatively wide and growing coverage (55 languages at the time of this writing). Our method could also potentially be used with other tools such as the Romanizer uroman[10], which is a less accurate phonetic representation than Epitran but covers most languages in the world. However, testing different romanizers is somewhat orthogonal to the main claims of this paper, and thus we have not explicitly performed experiments on this.

Our HRL pool contains 38 languages, specifically those that have more than 10k Wikipedia pages and are supported by Epitran. We do not consider Swedish and Cebuano because most Wikipedia pages of these two languages are bot-generated.[11]. We also remove all languages that do not achieve a candidate recall of 75% on the development set for the HRL, indicating that the model may not be trained well.

### 5.3 Main Results

Starting from the PBEL model, we gradually replace the baseline components with our proposed improvements to reach our complete model. The results are shown in the second section of Table 3. To put the results in the context, we also list the Wikipedia size and the hyperlink count of every language. While the Wikipedia size corresponds to the number of entities recorded in the Wikipedia, the hyperlink count roughly reflects the richness of the content of each page.

Overall, the model with the three proposed improvements yields significantly better performance than the baseline. It brings 7.4-33.3% improvement on top-30 gold candidate recall on six LRLs, with the exception of `te`. We will discuss the failure of `te` in §5.4.[12] Next, we can see that the CHARAGRAM brings the first major improvement, improving over both baselines BILSTM and CHARCNN. Even trained with $e_{\text{HRL}} - e_{\text{EN}}$ pairs, CHARAGRAM generalizes better to the test data ($m_{\text{LRL}} - e_{\text{EN}}$) where the patterns to be matched are different from the training data. This result suggests that, as we hypothesize, the model structure of CHARAGRAM makes it better able to learn string mappings in the face of relatively small and noisy data. We note that we also try many variations of the two baseline models. For example, we use the average hidden states instead of the last hidden state of BILSTM to represent a string, and we replace the sum-pooling layer with the max-pooling layer in CHARCNN. These variations yield comparable or worse recall compared to the current baselines.

In addition, introducing $m_{\text{HRL}} - e_{\text{EN}}$ pairs brings further improvement over all seven languages. This is perhaps not surprising; these data provide explicit supervision that matches the actual task of entity-mention matching that we are faced with at test time.

The influence of entity aliases varies from language to language. While they offer some significant gains in `om` and `mr`, they do not largely change other languages. We suspect this is due to the diverse properties of the languages used in our datasets. For example, for the case of Marathi speakers they may also speak English frequently and be familiar with English entity names due to English being a national language of India. This may lead them to follow conventions similar to the English aliases that are available in Wikidata. Speakers of other languages might either not use as many aliases or their aliases may not match well with those included in Wikidata.

Moreover, we quantify how our proposed methods reduce the failures existing in the baseline system. We use the 400 samples of §3 and compare the error distribution with the original one in Figure 4. From the results, we can see that our model eliminates a large number of the errors by ranking the correct entities the highest. It significantly reduces DIRECT and ALIAS errors, which demonstrates the effectiveness of our proposed method. As a side benefit, a number of the TRANS errors are also resolved. In addition, when the proposed model fails to rank the correct entity the highest, it is able to increase the number of correct entities in the top-30 candidate list, providing a downstream disambiguation model with larger improvement headroom. A few concrete examples are shown in Table 4.

---

[10] https://www.isi.edu/~ulf/uroman.html
[11] https://en.wikipedia.org/wiki/Lsjbot

[12] While not a direct target of our paper, we note that the three methodological improvements, especially the introduction of CHARAGRAM also improve the baseline model in HRL settings. We often observe more than a 20% gain in top-30 gold candidate recall in the development set, which is derived from the same HRL as the training set.

|  | DARPA-LRL | | | | WIKI | | | |
| Model | ti | om | rw | si | mr | lo | te | avg |
| --- | --- | --- | --- | --- | --- | --- | --- | --- |
| WikiMention | 21.9 | 45.3 | 59.6 | **66.6** | - | - | - | - |
| Translation | 13.4 | 20.9 | 25.3 | 21.0 | - | - | - | - |
| ee + BiLSTM = PBEL | 54.1 | 18.1 | 57.5 | 34.5 | 53.5 | 21.0 | **40.7** | 40.7 |
| ee + CharCNN | 53.8 | 13.0 | 55.9 | 30.8 | 47.7 | 18.0 | 24.6 | 34.8 |
| ee + Charagram | 70.6 | 20.4 | 60.2 | 17.5 | 63.4 | 40.1 | 23.8 | 43.2 |
| ee + me + Charagram | 74.4 | 41.3 | 64.6 | 50.7 | 72.8 | **54.4** | 34.3 | 56.6 |
| + aka = Ours | **75.1** | **46.0** | 64.9 | 51.1 | **77.5** | 54.3 | 34.4 | **57.6** |
| Wikipedia Size | 168 | 775 | 2K | 15K | 50K | 3K | 70K | 20K |
| Hyperlink Count | 188 | 4K | 7K | 63K | 300K | 11K | 610K | 165K |

Table 3: Top-30 gold candidate recall (%) of different models. First block: performance of direct Wikipedia-based models that use LRL resource; second block: performance of pivoting-based models that does not require any LRL resource. `ee` means using entity-entity pairs as training data and `me` means using mention-entity pairs as training data. **Bold** numbers are the best performance of the corresponding languages.

| Error Type | Mention | IPA | Ours | PBEL |
| --- | --- | --- | --- | --- |
| ALIAS | बीव्हर क्रीक स्की रिसॉर्ट<br>ग. दि. मा. | biːvɦərə kriːkə skiː risərʈə<br>gə. di. maː. | Beaver Creek Resort<br>Gajanan Digambar Madgulkar | Beaver Creek State Forest (New York)<br>Ghada Amer |
| DIRECT | हेर्मान स्टॉडिंजर<br>मुसोलिनीने | ɦermaːnə sʈəɖɪnd͡ʒər<br>musoliniːne | Hermann Staudinger<br>Muscoline | Herman Heuser<br>Benito Mussolini |
| TRANS | ख्मेर साम्राज्याचे<br>युरोपियन युनियनचे | kʰmerə saːmraːd͡ʒjaːt͡ɕe<br>juropijnə junijnət͡ɕe | Khmer Empire<br>European Union | Khmer Issarak<br>Yuri Petunin |

Table 4: Successful cases, where the top-1 candidate entity retrieved by our model improves over that of the baseline model.

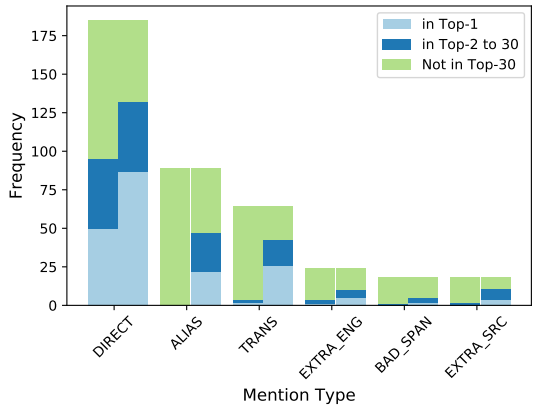

Figure 4: The distribution of mention types and the performance of our proposed model (right bars), compared with the baseline (left bars).

Up until this point, we have been comparing models that are purely zero-shot – they need no training data in the source LRL. However, even for low-resourced languages there is often *some* Wikipedia data that can be used to create models. Using this data, we additionally compare our model with the two Wikipedia-based models that are not zero-shot (§2.2) on four DARPA-LRL datasets on the first section of Table 3[13]. Our model consistently beats Translation on all four datasets without relying on any LRL resources. Moreover, it outperforms WikiMention by a large margin on three datasets with relatively small sized

---

[13]For the 3 WIKI datasets, the way we create these datasets is exactly the same as the way we generate $m_{\text{HRL}} - e_{\text{EN}}$ lookup tables, and thus WikiMention will achieve 100% recall. We skip the unfair comparison on these datasets.

| LRL | Linguistics | $n$-gram Overlap | $\delta$ |
|---|---|---|---|
| ti | âm, 63.9 (60.8) | am, 74.2 (70.9) | 10.3 |
| om | sô, 28.0 (63.7) | id̂, 40.9 (75.8) | 12.9 |
| rw | r̂n, 46.4 (62.9) | tl, 64.6 (79.0) | 18.2 |
| si | hi, 50.4 (63.1) | hi, 50.4 (63.1) | 0 |
| lo | th, 51.4 (78.8) | th, 51.4 (78.8) | 0 |
| mr | ĥi, 72.8 (83.3) | ĥi, 72.8 (83.3) | 0 |
| te | ta, 12.6 (32.3) | hi, 32.6 (45.1) | 20.0 |

Table 5: The pivoting language, performance (and their $n$-gram overlap % with the LRL) selected by different criteria. $\delta$ column shows the top-30 candidate recall improvement (%) using $n$-gram overlap. Language with a hat use grapheme representations while the remaining ones use phoneme representations.

Wikipedias, evidencing the advantage of zero-shot learning in resource scarce settings. For si with over 15K Wikipedia pages, our model lags behind the resource-heavy WIKIMENTION model by about 15% in the gold candidate recall. This is perhaps expected as our model does not rely on any of LRL resources, and it is possible that explicitly training our model with these resources could further improve its accuracy. Additionally, we observe that our model could serve as a complement to WIKIMENTION and bring further gain in gold candidate recall. We discuss this in detail in Section 5.6.

### 5.4 Pivoting Language Selection

Choosing a closely related HRL and directly applying the model trained on that HRL to the LRL has been a popular transfer learning paradigm in low-resource settings (Täckström et al., 2012; Zhang et al., 2016; Cotterell and Heigold, 2017; Rijhwani et al., 2019; Lin et al., 2019; Rahimi et al., 2019). Related languages are often chosen heuristically based on linguistic intuition, although there are some works that have recently examined training models to select languages automatically (Lin et al., 2019; Rahimi et al., 2019). In our case, we would like to choose both a pivoting language, and a string representation: phonemes or graphemes. This doubles the search space and increases the search difficulty.

We devise a simple yet strong heuristic for picking HRLs for transfer: picking the language that shares the largest number of character $n$-grams with the LRL. This is an automatic process that does not need any domain or linguistic knowledge. Table 5 shows the performance gap between this criterion and manual selection with linguistics features, which has been used in previous work on XEL (Rijhwani et al., 2019). Notably, to eliminate the variance caused by the different number of inter-language links possessed by different HRLs, we compare the similarity between $m_{\text{LRL}}$ with $e_{\text{EN}}$ directly, without the comparison between $m_{\text{LRL}}$ and $e_{\text{HRL}}$. More specifically, we replace Equation (1) with $\text{score}(m_{\text{LRL}}, e_{\text{EN}}) = \text{sim}(m_{\text{LRL}}, e_{\text{EN}})$.

It is clear that selecting proper pivoting languages and string representations is important; failing to do so can cause performance degradation of as much as 20%. However, while our heuristic selection method is empirically better than manual selection with linguistic features, it is notable that pivoting languages and the representations selected in this way do not necessarily yield the best performance. We observe that choosing a pivoting language with slightly less $n$-gram overlap yields better performance for some LRLs. For example, while om has about 43% character $n$-gram overlap with am, using the model trained with am yields a gold candidate recall of 45.0% (compared to 40.9% with id). This indicates that accuracy could be further improved with more sophisticated pivoting language selection criteria.

Regarding the importance of $n$-gram sharing, we suspect the relatively low recall of te compared to the baseline model results from a lack of shared character $n$-grams with its pivot language hi. While most other language pairs have over 60% character $n$-gram overlap, te and hi only have 45.1%, meaning $\mathbf{v}_m$ only encodes less than half $n$-grams it has. On the contrary, character-level embeddings used by BI-LSTM are less sparse than higher-order $n$-grams, and thus BI-LSTM suffers less information loss.

### 5.5 Properties of Learnt $n$-grams

As discussed in the previous sections, the objective of CHARAGRAM is to learn $n$-gram mappings between the HRL and English. To more concretely understand our model's behavior, we randomly sample a few English $n$-gram embeddings and retrieve their five nearest neighbors from the HRL side. Table 6 lists these most similar $n$-grams.

CHARAGRAM is able to correctly associate $n$-grams that have close pronunciation in different languages together. Because the pronunciation of the same syllable could vary in the context of different words, $n$-grams with small variances in vowels can still be reasonable approximations. For ex-ample, "li" can be pronounced as both "li" and "le" in different words. One thing that is worth mentioning is that CHARAGRAM is able to correctly recognize some mappings of non-transliterated words. For instance, "Jaamacadda" in so is the parallel of "University" in English, and the model was able to correctly align $n$-grams corresponding to these words. This result demonstrates one way how CHARAGRAM alleviates the TRANS error that BI-LSTM suffers from.

| HRL | EN | 5 Nearest Neighbor |
|---|---|---|
| am | ma | ma, marɨ, mo, \<s\>mo, \<s\>m |
|    | bi | bi, bija, bij, bɨja, əb |
| hi | ʃɑɹm | ʃərm, ʃərma, ʃərm, rmaː\</s\>, ʃər |
|    | lɪ | li, le, lin, laːi, aːli |
| th | lɪn | lin, lin\</s\>, lyn, lina, liːn |
|    | ʒejmz | ɕeːm, ɕeːm, ɕeːma, jame, mes |
| so | bi | bi, mbi, arbee, inho\</s\>, biya |
|    | Uni | maca, amac, Jaam, macad, \<s\>Jaam |

Table 6: Randomly sampled English n-grams and their five nearest neighbors in n-gram embedding space.

|    | ee + BiLSTM | Ours | δ |
|---|---|---|---|
| ti | 50.8 (55.4) | 67.5 (75.8) | 16.7 (20.4) |
| om | 53.2 (61.3) | 59.2 (67.9) | 6.0 (6.6) |
| rw | 61.5 (67.5) | 68.9 (73.9) | 7.4 (6.4) |
| si | 70.9 (76.1) | 72.2 (78.0) | 1.3 (1.9) |
| avg | 59.1 (65.1) | 67.0 (73.9) | 7.9 (8.8) |

Table 7: In-KB accuracy (with top-30 gold candidate recall of the merged candidate lists in brackets, both represent percentage %) of the end-to-end XEL system with different candidate generation models. $\delta$ shows the in-KB accuracy degrade (%) using baseline candidate generation model.[16]

### 5.6 Improving End-to-end XEL Systems

To investigate how our candidate generation model influences the end-to-end XEL system, we use its candidate lists in the disambiguation model BURN proposed by Zhou et al. (2019). BURN creates a fully connected graph for each document and performs joint inference on all mentions in the document. To the best of our knowledge, it is currently the disambiguation model that has demonstrated the strongest empirical results for XEL without any targeted LRL resources. Therefore, we believe it is the most reasonable choice in our low-resource scenario. For details, we encourage readers to refer to the original paper.[14]

To make the best use of scarce but existing resources, we follow Zhou et al. (2019) and concatenate candidate lists generated by WIKIMENTION to candidate lists of both the baseline and our method. The score of each candidate entity is calculated in the following way:

$$\text{score}_{\text{merge}}(e_{\text{EN}}) = \alpha \times \text{score}_{\text{wm}}(e_{\text{EN}}) + (1-\alpha) \times \text{score}'_{\text{ca}}(e_{\text{EN}})$$
$$\text{score}'_{\text{ca}}(e_{\text{EN}}) = \text{softmax}(\beta \times \text{score}_{\text{ca}}(e_{\text{EN}}))$$

where $\text{score}_{\text{wm}}$ is the score from WIKIMENTION and $\text{score}_{\text{cn}}$ is the original score from CHARAGRAM. $\text{score}'_{\text{cn}}$ is the scaled score over the top-30 candidate list. We omit $m_{\text{LRL}}$ in all score functions for simpilicty. In our experiments, $\alpha$ is set to 0.6 and $\beta$ is set to 100.

Table 7 lists the end-to-end XEL results. Compared to the baseline model, our model recovers more candidate entities missed by WIKIMENTION and significantly benefits the downstream disambiguation model, as well as the end-to-end system. Even though incorporating WIKIMENTION narrows the gap of gold candidate recall (compared to Table 3), our model still beats the baseline model by a large margin. While the baseline candidate generation model only reaches a recall in the range of 60% on average, ours yields a recall in the range of 70%, closer to the high-resource counterparts which are often in the range of 80%. As a result, the end-to-end XEL in-KB accuracy increases over all four languages, with gains from 1.3 to 16.7%. This is significant for extremely low-resource languages like ti, indicating the potential of our model in truly resource-scarce settings.

## 6 Related Work

**Candidate generation for entity linking**: In most work, candidate generation for monolingual entity linking relies on string matching and Wikipedia anchor text lookup (Shen et al., 2015). For cross-lingual entity linking, inter-language links from Wikipedia and bilingual lexicons are used to translate the given entity mentions into the language of the KB (often English) in order to generate candidates (Tsai and Roth, 2016; Pan et al., 2017; Upadhyay et al., 2018a). More recently, Rijhwani et al. (2019) use orthographic and phonological similarity to high-resource languages to generate candidates for low-resource test languages. For the related task of clustering entities, Blissett and Ji (2019) use RNNs for measuring orthographic similarity of entity mentions.

---

[14]It is notable that we assume that the XEL system could access the oracle NER outputs. In reality, the F1 scores of low-resource NER are often in the range of 70%. We leave the evaluation and possible improvement with non-perfect NER systems as our future work.

[16]These results are not comparable to Rijhwani et al. (2019) as we only consider a subset of mentions whose linked entity exists in the Wikipedia.

**Transliteration**: There has also been work in transliterating named entities from one language to another (Knight and Graehl, 1998; Li et al., 2004). Although similar to our current task of selecting candidates from an English KB, transliteration poses different challenges as it involves *generating* the English entity name itself. Upadhyay et al. (2018b) use a sequence-to-sequence model and a bootstrapping method to transliterate low-resource entity mentions using extremely limited training data. (Tsai and Roth, 2018) combine the standard translation method for XEL candidate generation with a transliteration score to improve XEL candidate recall on several languages.

**Bilingual lexicon induction**: Another related task is bilingual lexicon induction, where a mapping between words in two languages is predicted by a learned model (Haghighi et al., 2008). Although such a mapping can be used to translate entities from the source test language to English for XEL candidate generation, most existing lexicon induction methods assume the availability of a large amount of monolingual data in both the source and target language (Conneau et al., 2017; Chen and Cardie, 2018; Artetxe et al., 2018). Although this data is readily available in English, it is unrealistic for many low-resource languages, diminishing the utility of such methods for the low-resource XEL task.

## 7 Conclusion

In this work, we perform a systematic analysis to study and address the limitation of a previous candidate generation model in low-resource settings. We propose three methodological improvements to resolve two main problems of the baseline model, namely, mismatch between mention and entity and sub-optimal string modeling. For the first problem, we introduce mention-entity pairs into the training process to provide supervision. We additionally collect entity aliases from English Wikidata to further bridge this gap. To solve the second problem, we replace the LSTM with a more direct model CHARAGRAM. These methods form our proposed candidate generation model. We experiment with seven realistic datasets in LRLs. Our model yields an average gain of 16.9% in top-30 gold candidate recall. We also evaluate the influence of our candidate generation model in the context of end-to-end low-resource XEL. It brings an average gain of 7.9% in four LRLs.

An immediate future focus is finding a way to properly combine multiple models trained on different HRLs together to have better character n-gram coverage and thus improve model performance in different LRLs. Another interesting avenue is to investigate how to efficiently compare mentions and a large number of entities (e.g., 2M in Wikipedia) in high dimensional space. Currently, our model calculates the cosine similarity between a mention and every entity in the KB, which takes a few minutes for each test set. However, there is much existing work (Rajaraman and Ullman, 2011; Johnson et al., 2019) for efficient similarity search in high dimensional space for billion-scale datasets. It is likely that combination of these algorithms with our retrieval method will allow them to scale well and reduce the computation time to a few seconds. In addition, other interesting future directions are examining how to balance the trade-off between the gold candidate recall and the disambiguation difficulty, and how to apply our model to settings where the target language is not English.

## 8 Acknowledgements

We would like to thank Radu Florian and the anonymous reviewers for their useful feedback. This material is based upon work supported in part by the Defense Advanced Research Projects Agency Information Innovation Office (I2O) Low Resource Languages for Emergent Incidents (LORELEI) program under Contract No. HR0011-15-C0114. The views and conclusions contained in this document are those of the authors and should not be interpreted as representing the official policies, either expressed or implied, of the U.S. Government. The U.S. Government is authorized to reproduce and distribute reprints for Government purposes notwithstanding any copyright notation here on. Shruti Rijhwani is supported by a Bloomberg Data Science Ph.D. Fellowship.